\documentclass[letterpaper, 10 pt, journal, twoside]{IEEEtran}
\usepackage[utf8]{inputenc}

\usepackage{graphicx} 
\usepackage{adjustbox}

\usepackage{booktabs} 
\usepackage{array} 
\usepackage{paralist} 
\usepackage{verbatim} 
\usepackage{subfig} 
\usepackage{amsmath}
\usepackage{amssymb}
\usepackage{mathdots}
\usepackage[ruled,vlined]{algorithm2e}
\usepackage{todonotes}
\usepackage{bm}
\usepackage{xargs}
\usepackage{comment}
\usepackage{algorithmic}
\usepackage{standalone}
\usepackage{tikz}
\usetikzlibrary{decorations.pathreplacing}
\usetikzlibrary{automata,positioning}
\usepackage{tikz-cd}
\usepackage{tikz-network}

\newtheorem{thm}{Theorem}[section] 

\newtheorem{defn}[thm]{Definition} 
\newtheorem{lem}[thm]{Lemma}
\newtheorem{corol}[thm]{Corollary}

\newcommand{\Act}{\mathcal{A}}
\newcommand{\Inp}{\Sigma}
\newcommand{\equal}{=}

\newcommand\copyrighttext{%
  \footnotesize \textcopyright 2021 IEEE. Personal use of this material is permitted.  Permission from IEEE must be obtained for all other uses, in any current or future media, including reprinting/republishing this material for advertising or promotional purposes, creating new collective works, for resale or redistribution to servers or lists, or reuse of any copyrighted component of this work in other works.}
\newcommand\copyrightnotice{%
\begin{tikzpicture}[remember picture,overlay]
\node[anchor=south,yshift=10pt] at (current page.south) {\fbox{\parbox{\dimexpr\textwidth-\fboxsep-\fboxrule\relax}{\copyrighttext}}};
\end{tikzpicture}%
}

\tikzset{commutative diagrams/.cd,
mysymbol/.style={start anchor=center,end anchor=center,draw=none}
}

\DeclareMathOperator*{\seq}{\rightarrow}
\DeclareMathOperator*{\fb}{?}

\begin{document}
\title{An expressiveness hierarchy of Behavior Trees and related architectures}
\author{Oliver Biggar$^{1}$, Mohammad Zamani$^{1}$, and Iman Shames$^{2}$
\thanks{Manuscript received: December 17, 2020; Revised February 12, 2021; Accepted April 6, 2021.}
\thanks{This paper was recommended for publication by Editor Markus Vincze upon evaluation of the Associate Editor and Reviewers' comments. This  work  was  funded  by   Defence  Science  and  Technology  Group,  through  agreement MyIP: 10266 entitled ``Hierarchical Verification of Autonomy Architectures'', and the Australian Government, via grant AUSMURIB000001 associated with ONR MURI grant N00014-19-1-2571.}
\thanks{$^{1}$Oliver Biggar and Mohammad Zamani are with the Defence Science and Technology Group, Australia
        {\tt\footnotesize oliver.biggar1@dst.defence.gov.au}, {\tt\footnotesize mohammad.zamani@dst.defence.gov.au}}%
\thanks{$^{2} $Iman Shames is with School of Engineering, Australian National University, Australia
        {\tt\footnotesize iman.shames@anu.edu.au}}%
\thanks{Digital Object Identifier (DOI): see top of this page.}
}

\maketitle
\copyrightnotice
\begin{abstract}
In this paper we provide a formal framework for comparing the expressive power of Behavior Trees (BTs) to other action selection architectures. Taking inspiration from the analogous comparisons of structural programming methodologies, we formalise the concept of `expressiveness'. This leads us to an expressiveness hierarchy of control architectures, which includes BTs, Decision Trees (DTs), Teleo-reactive Programs (TRs) and Finite State Machines (FSMs). By distinguishing between BTs with auxiliary variables and those without, we demonstrate the existence of a trade-off in BT design between \emph{readability} and \emph{expressiveness}. We discuss what this means for BTs in practice.
\end{abstract}


\section{Introduction}

\IEEEPARstart{D}{iscrete} decision-making or task-switching is an important component of much high-level artificial intelligence in robotics and autonomous systems applications. Such decision-making can be represented in various forms, such as Finite State Machines, Teleo-reactive programs, Decision Trees and Behavior Trees. We refer to these concepts collectively as `control architectures'~\cite{btbook}, a term we shall use to mean a formal representation of discrete decision-making.

In this paper we build a framework to analyse control architectures in terms of their expressiveness, which we define as the breadth of behaviors that they can theoretically construct. Moreover, we examine the trade-offs  between expressiveness and readability. Our goal is to determine the suitability of different control architectures for different applications. 
A useful source of intuition is the programming literature, in which there has been much work done to answer similar questions.

Almost all practical programming languages are Turing-complete, meaning that they can express all possible algorithms (when one allows computers arbitrary amounts of memory). As a result, expressiveness is not always sufficient to decide which programming language to use for a particular application. In 1968, Edsger Dijkstra wrote a letter~\cite{gotoharmful} which argued that structuring programs in code blocks organised by simple flow control which avoided the \textbf{go to} statement led to more reusable and readable code. This argued that these properties were important for comparing programming languages. This led to a transformation, known as the \emph{structured programming} movement, which has allowed today's programs to be orders of magnitude larger in size and complexity while still being maintainable.

The benefits of readability and reusability are receiving growing interest in AI and robotics~\cite{brooks1986robust,nilsson1993teleo}, because these properties are useful for control architectures just as they are for programs. As large-scale autonomous systems become widespread, there is growing appreciation of architectures which can be stored, extended and recombined to build agent behavior of great complexity~\cite{btbook,surveyofbts,integratedARM,understandinggameAI}.

In arguing why \textbf{go to} statements are harmful in programming, Dijkstra~\cite{gotoharmful} states that ``our powers to visualize processes evolving in time are relatively poorly developed" and thus we should endeavour to make the ``correspondence between the program (spread out in text space) and the process (spread out in time) as trivial as possible.". The same is true in robotics, where we wish to ensure a close correspondence between the structure of a control architecture and the behavior of a robot, to allow understanding, reuse and debugging. We shall use this as our definition for \emph{readability}: a control architecture is \emph{readable} if there is a close correlation between its visual structure and the behavior it induces in a robot.

In the past, Finite State Machines (FSMs) have been dominant in AI, despite the fact they increase in fragility with size~\cite{ogren2012increasing,bojic2011extending}. Specifically, as more states and arcs are added, it becomes harder to track and understand the \emph{paths} through an FSM~\cite{btbook}, reducing its readability. Recently, a tree-structured control architecture called the Behavior Tree (BT) has been growing in popularity, with proponents arguing it could be a replacement for FSMs. In fact, a number have argued FSMs resemble unconstrained use of the \textbf{go to} statement, and that BTs instead resemble structured programs~\cite{ogren2012increasing,generalise,btbook}. The tree structure of BTs and their left-to-right execution\footnote{See Section~\ref{sec:bts} for a definition of BTs.} makes their structure closely resemble the execution they induce, so at any instant it is straightforward to determine which action is selected. BTs have two critical properties: \emph{modularity}, in that every subtree is itself a BT, which means that subtrees can be read as sub-behaviors; and \emph{reactiveness}, in that BTs frequently select a new behavior independently of the previous, and so are responsive to changes in the environment. 
These properties make BTs \emph{readable}~\cite{ogren2012increasing,surveyofbts,generalise}, allowing them to be easily designed and understood, and have been suggested~\cite{btbook,generalise,surveyofbts,UAVmissionBT,kbts} as the reason why BTs are reportedly of significant interest in robotic applications~\cite{integratedARM,hu2015semi}.


To understand when the use of a BT might improve our designs, we must know if the behavior we desire can be expressed by a BT. We extend and formalise the existing work on this problem into a single hierarchy of expressiveness, which includes BTs, FSMs, Decision Trees~\cite{btbook}, and Teleo-reactive Programs~\cite{nilsson1993teleo}. In the process we distinguish between versions of these architectures which can read and write to auxiliary variables, and those that cannot, and discuss the result from~\cite{ogren2012increasing} that BTs with this capability are as expressive as FSMs. We strengthen this result to show that using auxiliary variables allows all control architectures to be proven equally expressive.

Some programmers interpreted Dijkstra's letter as an argument to completely abolish the \textbf{go to} statement and similar jump-like constructions. This was possible~\cite{bohm1966flow}, at least in theory. However, Knuth~\cite{knuth1974structured} later pointed out that while removing \textbf{go to} statements usually led to clearer code, there were a few cases where a jump statement remained the most efficient and readable choice. In other words, \textbf{go to}-less programs were not always `better', because occasionally avoiding a \textbf{go to} was more convoluted than using one explicitly. Knuth addressed this problem by outlining where a \textbf{go to} made sense and where it should be avoided. In a similar way, we argue in Section~\ref{sec:expressivenes} that while BTs may often lead to clear designs, manipulating auxiliary variables within a BT is not especially readable, and so can counteract the benefits of BTs. We show that auxiliary variables (or some form of accessible information storage) is \emph{necessary} to achieve significant expressiveness. Specifically, we cannot in general construct a BT to emulate an FSM using only the actions and conditions to which that FSM has access. We conclude that there is a trade-off in BT design between readability and expressiveness.
Just as many modern programming languages occasionally violate the strictures of structured programming (usually through the use of \textbf{break} and \textbf{return} statements), it is likely the case that FSMs or FSM-like constructions cannot always be avoided. Understanding and appropriately handling such cases is key to pursuing readable robotic intelligence. 

 The rest of the paper is organised as follows. Section~\ref{sec:related work} describes the existing research in this area. Section~\ref{sec:asms} introduces our formal definition of expressiveness, our assumptions, and explains how a number of commonly-used control architectures are interpreted in this framework. Our comparisons and findings are in Section~\ref{sec:expressivenes}. Finally, Section~\ref{sec:conclusions} concludes the paper.
 
 \section{Related Work} \label{sec:related work}
 
 

Our paper follows an approach similar to that of Kosaraju~\cite{kosaraju1974analysis}, who studied the problem of comparing different structured programming paradigms. We discuss more deeply the oft-stated analogy between BTs and function calls in structured programs~\cite{ogren2012increasing,generalise,btbook}, and analyse at length the previous result of~\cite{ogren2012increasing} equating BTs and FSMs in the context of the \emph{structured program theorem}~\cite{bohm1966flow}.

Colledanchise and \"Ogren~\cite{colledanchise2016behavior,generalise} have similarly explored relationships between control architectures, particularly in comparison to BTs. They show that BTs `generalise' TRs, DTs, FSMs, sequential behavior compositions and And-Or Trees. Our definition of expressiveness is largely consistent with these proofs, and we make use of their results. We also show that using their assumptions (where auxiliary variables can be used, and  which correspond to what we call \emph{unrestricted} architectures in Section~\ref{sec:expressivenes}) the opposite inclusions are possible, so all these architectures are equally expressive. We then extend these results by showing the consequences of forbidding auxiliary variables, and then discuss the results, particularly relating to a proof from~\cite{ogren2012increasing}, which shows unrestricted BTs are as expressive as FSMs.

\section{Formalising architectures as action selection mechanisms}\label{sec:asms}

Our aim in this paper is to compare the expressiveness of BTs and other discrete task-switching architectures, which we refer to as control architectures. To formalise this comparison, we provide a mathematical definition of their input-output behavior, namely, their  \emph{action selection mechanism}. In Section~\ref{sec:execution mode} we explain the assumptions behind this definition, and how a number of control architectures can be modeled in this way. Later in Section~\ref{sec:expressivenes} we apply this definition and provide an `expressiveness' hierarchy. 

\begin{defn} \label{def:asm}
Let $\Act$ be a finite set of \emph{actions} and $\Inp$ a finite set of \emph{inputs}. An \emph{action selection mechanism} (ASM) is a map $f:\Inp^*\to \Act$, where $\Inp^*$ is the set of all finite sequences of elements of $\Inp$. In other words, an ASM takes a finite sequence of inputs and produces an action.
\end{defn}

Actions are abstract descriptions of physical tasks that the robotic system can perform, such as `Open door'. Inputs are an abstract representation of the physical environment to which the robot responds, such as the proposition `door open'. We explain this at greater length in the next section, but in general we treat the actions in $\Act$ and inputs in $\Inp$ as black boxes. ASMs are mappings, so to compare them we fix their domain and codomain, which unless otherwise specified is assumed to be $\Inp$ and $\Act$ respectively for the ASMs in this paper.


\subsection{The execution model of ASMs} \label{sec:execution mode}

In this section we shall explain how this abstract definition of an ASM applies concretely in robotics.

Consider a robotic system, whose discrete task-switching is controlled by an ASM $f$, which selects from among its finite actions or tasks $\Act$. While this system gathers significant data about itself and its surroundings via its sensors, this is not necessarily what we consider to be input $\Inp$ to its discrete task-switching architecture. Instead, we shall define $\Inp$ to be the finite set of abstracted information the ASM can use to decide between actions. For instance, if there is a finite set $K$ of Boolean variables used to make task selections then $\Inp=2^K$, where each $x\in\Inp$ is the set of variables true at that point. Thus if we have a BT controlling a robot, we do not necessarily consider the sensor data directly as input to the BT. Instead, input is considered as combinations of propositions of the form `Condition $C$ returns Success' and `Action $A$ returns Failure'. These propositions may well be based on processed sensor data (such as `Wall Ahead'), but the processing itself is not the concern of the BT considered as a discrete task-switching architecture. Therefore in this setup, a BT is to only use the return values of its Conditions and Actions to decide on its selection decision; if at any two instants these values are the same, the same action is selected.

Treating the input in an abstract form isolates the computations performed at the discrete task-switching layer. To construct this abstract input, computation must occur; while we expect this, it is not the concern of the ASM as defined in this paper. For instance, one typically employs computer vision systems, state estimators, machine learning algorithms and other crucial parts of a real-world robotic system to be responsible for low-level data processing.  It should also be noted that the abstractly represented input is often subject to processing error and cannot be assumed as `true' representation of the `real' world. ASMs, such as BTs and FSMs, merely query the abstractly represented input information in a Boolean form and make a decision on that basis.

The fact that ASMs take a finite \emph{sequence} of input (See Definition~\ref{def:asm}) represents progress in time. That is, the first input $x_0\in\Sigma$ is given at the beginning of the robot's execution, and is used to select its first action $f(x_0)$. At the next decision instance (which we assume occur frequently in real-world time; these corresponds to `ticks' in the usual BT terminology\footnote{The `frequent-ticks' assumption here covers the fact that, consistent with BT literature~\cite{btbook}, we do not wait for Actions to `complete' before a return value is given to the BT. Consider an action Push Block, which returns Failure if the block is too heavy. Two robots in seemingly identical circumstances may decide to attempt Push Block but later have one Succeed and the other Fail. With this assumption, this situation is modeled as follows. At the initial decision instant the input is identical and the same decision is made (Push Block), which in both cases returns Running. Only at a \emph{later} time step (and thus a different input) do we find the states to be distinguishable (by say, more force being registered by sensors, causing the action in that BT to return Failure), and the corresponding ASM selection may differ.}), $x_1$ is provided to the ASM, which selects action $f(x_0,x_1)$, and so on. 
From an operational perspective the ASM merely takes the values of some atomic propositions at every update step and produces an action in response. Allowing $f$ to use the entire finite sequence of input (which consists of all inputs provided to the system up until that time) represents that the decision-making may use internal state. Thus even though a single input is provided at any time, $f$ is allowed to potentially make use of past inputs to influence its current selection. In an FSM, for example, past inputs affect the current state of the machine, which in term affects responses to future inputs. Overall then, an ASM responds to sequences of input (from $\Inp$) with sequences of actions (from $\Act$).
\begin{align*}
\text{Input:}&\ x_0,x_1,x_2\dots \\ \text{Output:}&\ f(x_0),f(x_0,x_1),f(x_0,x_1,x_2),\dots
\end{align*}
These sequences of actions we call \emph{behaviors}. 



\subsection{Behavior Trees} \label{sec:bts}
Behavior Trees (BTs)~\cite{btbook} are control architectures structured as ordered directed trees. The execution of a BT occurs through signals called `ticks', which are repeatedly generated by the root node and sent to its children in the left-to-right order. A node is executed when it receives ticks. All nodes return one of three possible values when ticked: Success, Failure or Running. Leaf nodes (called \emph{execution nodes}) come in two types, Condition and Action. A Condition (drawn as an ellipse) checks some proposition, returning Success if true and Failure otherwise. An Action (drawn as a rectangle) represents an action taken by the agent, Actions return `Success' if they have achieved their goal, `Failure' if they cannot operate and `Running' otherwise, indicating execution is underway. Internal nodes (called \emph{control flow nodes}), when executed, tick their children from left to right, and come in four types (Sequence, Fallback, Parallel, and Decorator). Sequence nodes (drawn as a $\seq$ symbol) tick their children in order, and if any return Failure or Running that value is immediately returned by the Sequence node, and Success is returned only if every child returns Success. Fallback (drawn as a $\fb$) is analogous to Sequence, except that it returns Failure only if every child returns Failure, and Success or Running if any child returns those. We shall omit discussion of the Parallel node in this paper, to avoid a discussion of concurrency. The Decorator node returns a value based on any user-defined policy regarding the return values of its children. For more detail, see~\cite{btbook}.
 
  
  When interpreting BTs as ASMs, the Action nodes are the actions, and the conditions for returning Success, Failure and Running are formulated on $K$ (the finite set of variables used for discrete task-switching). On each input $x_i\in\Sigma$, we tick (execute) the tree starting from the root, as described above, and select the leaf node that was ticked last in the traversal of the tree. If any leaf node returns Running when ticked, then the root returns Running, making that node (action) the last one ticked and thus the one selected by the tree.\footnote{
   BTs return the last leaf ticked, which may not necessarily return Running on that input. This definition is still reasonable, because the root returns Success/Failure/Running if and only if the last child ticked returns that value (with the further assumption that Negation decorators are not used. By Lemma 4.18 of~\cite{biggar2020framework} all Negation decorators can be equivalently propagated downwards and absorbed into modified leaves). The overall return value of a BT can therefore be derived from its selected action at any step~\cite{modularity}. 
    A different approach~\cite{hannaford2019hidden} to representing `overall tree Success/Failure' is to model these as additional trivial actions which can be selected by the tree, but this does not integrate with composition. A BT which selects a `Success' Action instead of returning Success can't be embedded properly as a subtree of another BT.
   }

\subsection{Teleo-reactive programs and Decision Trees}
Teleo-reactive programs (TRs)~\cite{nilsson1993teleo} are lists of condition-action rules,\begin{equation*}
    k_1 \to a_1,\ 
    k_2 \to a_2,\  
    \dots,\ 
    k_n \to a_n, 
\end{equation*} that are executed by continuously scanning the list of conditions $k_i$ (formulated on $K$) in the order they appear in the list, and selecting the action $a_i\in\mathcal{A}$ associated with the first satisfied condition. As these conditions become true and false, the selected action changes immediately to the action corresponding to the first satisfied condition. The \emph{teleo} indicates that such lists are goal-oriented while \emph{reactive} is intended to describe how they react constantly to changes in the environment. This notion of reactiveness agrees with the definition in terms of ASMs we give later. As ASMs, TRs process each input $x_i$ from the start of the list, test in order whether any preconditions are true in $x_i$, and return the action corresponding to the first true precondition.

 Decision Trees (DTs)~\cite{btbook} are decision-making tools structured as binary trees where leaves represent actions and internal nodes represent predicates. The two arcs out of each predicate are labelled by `True' and `False'. Execution of DTs occurs by beginning at the root and evaluating each predicate on the current input state, and proceeding down the `True' or `False' arc depending on its value. When a leaf is reached, the action labelling that leaf is executed. 
As an ASM, the exact same process occurs on a given input, and the selected action is returned. Like TRs, DTs are executed reactively by repeatedly checking the tree starting from its root against the current state of the world.


\subsection{Finite State Machines}
 
A Finite State Machine (FSM) is a control architecture structured as a labelled directed graph, where nodes represent \emph{states}, each of which is labelled by the action performed in that state. An example is Fig.~\ref{fig:fsm counterexample}. The FSM is in exactly one of these states at any given time, and has one state as the initial state. Arcs, called \emph{transitions}, link states from one to another. A transition from one state to another is undertaken in response to input which \emph{triggers} the transition. FSMs begin in the initial state, and whenever input is received transitions are followed which are triggered by that input, until a state is reached where no more transitions are triggered. The FSM executes the action labelling the state until new input is received, at which point it continues from that state. Formally, an FSM is a six-tuple $(Q,q_0,\Inp,\Act,\delta,\ell)$, where $Q$ is a finite set of states, $q_0\in Q$ is the initial state, $\Inp$ is a finite set of inputs, $\Act$ is a finite set of actions, $\delta:\Inp\times Q\to Q$ is the transition function dictating when transitions are triggered and $\ell:Q\to \Act$ is the output function assigning states to actions. This particular formal interpretation of a Finite State Machine is called a \emph{Moore machine}~\cite{moore1956gedanken}.

 To consider an FSM as an ASM, given a sequence of input $x_0,\dots,x_n\in\Inp^*$, we begin at $q_0$ and apply the inputs in order. For each $x_i$, if a transition is triggered in the current state we update the current state to the head of the transition and repeat until we reach a state $q$ from which $x_i$ triggers no transitions. We assume that such a state exists and is reached after finitely many transitions. If there is no more input ($i=n$), return the action $\alpha$ labelling $q$, otherwise process the next input $x_{i+1}$, starting at $q$.

\subsection{Reactiveness} 

BTs, TRs, and DTs share a core similarity as ASMs. On an input sequence $x_0,\dots,x_n$, only the final input $x_n$ has an impact on the selected action. This means these architectures select the current action on the basis of the current input alone. This is a core element of their execution model and we shall use this to inspire the definition of a \emph{reactive architecture}. 

\begin{defn}\label{def:reactive}
A \emph{reactive architecture} is one whose ASM $f:\Inp^*\to \Act$ can be expressed by a single map $r:\Inp\to\Act$, so $f(x_0\dots x_n)= r(x_n)$.
\end{defn}
Reactive architectures such as BTs, TRs and DTs restart their execution from their root on each new input. By contrast, FSMs are not reactive (in the sense of Definition~\ref{def:reactive}) as they do not reset from their initial state on a new input, but instead continue from the current state, which depends on prior inputs. The final action selected is thus not solely dependent on the final input $x_n$. Observe that to define the ASMs of reactive architectures it was sufficient to describe the selection for a single input. Reactive architectures are often readable, because their sequence of selections are essentially independent, and so visualising the overall robotic behavior is no more difficult than visualising the selection process for a single action.

Note also that reactiveness in Definition~\ref{def:reactive} is defined with respect to a given set of inputs $\Inp$. Consider FSMs. The action selected by an FSM can be represented by the map $\ell\circ\delta:\Inp\times Q\to \Act$. If $\Inp\times Q$ was considered as the input set, then this map defined a reactive ASM. However, $\Inp$ is fixed, so that we have a common input on which to compare architectures, so we do not consider FSMs to be reactive\footnote{Judging by Definition~\ref{def:reactive}, other FSM-like structures such as Markov Decision Processes are likewise not reactive.}.

\section{Expressiveness} \label{sec:expressivenes}

In this section we explore the \emph{expressiveness} of BTs and other architectures, that is, whether we can construct certain behavior using a given set of actions. We begin by defining architectures which can and cannot access auxiliary variables, and then we present a theorem which groups a number of control architectures into a hierarchy of expressiveness.

\begin{defn}
Let $X$ be a class of control architectures, such as BTs or FSMs. Let $v$ denote an auxiliary integer variable  which can store $n$ different values $i\in\{1,\dots,n\}$. Let $Y$ denote the set of actions of the form `Write $i$ to $v$' and let $P$ denote the set of propositions of the form `$v=i$ ?' for any $i\in\{1,\dots,n\}$. Recalling our usual input and action sets $\Inp=2^K$ and $\Act$, we define a \emph{pure X} as an ASM $f$ from an $X$ with $f:\Inp^*\to\Act$. Then, we define an \emph{unrestricted X} as an ASM $g$ from an $X$ with $g:(2^K\times P)^*\to(\Act\times Y)$. Conceptually, unrestricted architectures can make their decision on the basis of both the variables in $K$ and the value of the auxiliary variable $v$, and can modify the value of $v$ while performing an action from $\Act$. When referring to the ASMs from unrestricted and pure forms of various architectures, we shall write \textbf{uX} and \textbf{pX}. So for example, \textbf{pBT} and \textbf{uFSM} denote the ASMs from pure BTs and unrestricted FSMs respectively\footnote{We cannot immediately compare pure and unrestricted architectures because their domains and codomains differ. We avoid this problem formally by considering a pure ASM $f$ and unrestricted ASM $g$ equivalent if $\pi_1\circ g = f\circ \pi_1^*$, where $\pi_1:(\Act\times Y)\to \Act$ is the projection onto $\Act$, and $\pi_1^*:(2^K\times P)^*\to (2^K)^*$ is the projection onto the sequences of input. For instance, when comparing the unrestricted BT in Fig.~\ref{fig:fsm_bt} to a pure FSM, we shall consider the pair of actions `Write 1 to $v$; do Action $a$' to be the same as an action `do Action $a$' in the pure FSM. We do not dwell on this point however, because it is only needed for the proofs of \textbf{uFSM} = \textbf{pFSM} (which we largely omit) and \textbf{uBT} = \textbf{pFSM} (which has already been proved in~\cite{ogren2012increasing}, where this point was not discussed). Further, the complication caused by this supports our conclusions of Section~\ref{sec:bts vs fsms} and~\ref{sec:conditions} that not treating actions as black boxes reduces readability. 
}.
\end{defn}
\begin{defn}
Let $\text{\textbf{X}}$ and $\text{\textbf{Z}}$ be two classes of ASMs. We say $\text{\textbf{X}}$ is \emph{more expressive than} $\text{\textbf{Z}}$ if $\text{\textbf{Z}}\subset \text{\textbf{X}}$, and \emph{equally expressive} if $\text{\textbf{Z}} = \text{\textbf{X}}$.
\end{defn}

We shall also write \textbf{Reactive} for the set of all reactive ASMs with input from $2^K$ and actions $\Act$. This leads us to Theorem~\ref{asm hierarchy}, which presents results from a variety of sources in a single hierarchy. 
Before stating the main result we state the following lemma. This lemma is a modified version of a similar result from~\cite{ogren2012increasing}. Here, the memory use is made explicit, and the `Write' and `do Action' are merged. We separate Lemma~\ref{implict memory+bt=fsm} and Fig.~\ref{fig:fsm_bt} because we discuss them specifically in Section~\ref{sec:bts vs fsms}.
\begin{figure}
    \centering
    \begin{tikzpicture}
\clip (0,0) rectangle (8.0,4.0);
\Vertex[x=4.045,y=3.600,size=0.4,opacity=0,label=?,fontscale=1.2,shape=rectangle,style={}]{1}
\Vertex[x=4.045,y=2.9,size=0.4,opacity=0,label=$\rightarrow$,fontscale=1.2,shape=rectangle,style={}]{2}
\Vertex[x=1.303,y=2,size=0.7,opacity=0,label=if $v\equal{} i$,fontscale=1.2,shape=ellipse,style={minimum width = width("fSM in Sta")+2ex}]{3}
\Vertex[x=5.3,y=2.2,size=0.4,opacity=0,label=?,fontscale=1.2,shape=rectangle,style={}]{4}
\Vertex[Pseudo,x=5.455,y=3,size=0.7,opacity=0,label=$\dots$,fontscale=1.2,shape=ellipse,style={minimum width = width("any")+2ex}]{6}
\Vertex[Pseudo,x=2.555,y=3,size=0.7,opacity=0,label=$\dots$,fontscale=1.2,shape=ellipse,style={minimum width = width("any")+2ex}]{7}
\Vertex[x=3.6,y=1.6,size=0.4,opacity=0,label=$\rightarrow$,fontscale=1.2,shape=rectangle,style={}]{8}
\Vertex[Pseudo,x=6.9,y=1.6,size=0.4,opacity=0,label=$\dots$,fontscale=1.2,shape=rectangle,style={}]{9}
\Vertex[x=1.9,y=0.4,size=0.7,opacity=0,label=Transition condition 1,fontscale=1.2,shape=ellipse,style={minimum width = width("Transition condition ")+2ex}]{10}
\Vertex[x=5.8,y=0.4,size=0.7,opacity=0,label=Write 1 to $v$; do Action $i$,fontscale=1.2,shape=rectangle,style={minimum width = width("Write 1 to v do action 1")+2ex}]{11}
\Edge[,fontscale=1.2,Direct](1)(2)
\Edge[,fontscale=1.2,Direct](1)(6)
\Edge[,fontscale=1.2,Direct](1)(7)
\Edge[,fontscale=1.2,Direct](2)(3)
\Edge[,fontscale=1.2,Direct](2)(4)
\Edge[,fontscale=1.2,Direct](4)(8)
\Edge[,fontscale=1.2,Direct](4)(9)
\Edge[,fontscale=1.2,Direct](8)(10)
\Edge[,fontscale=1.2,Direct](8)(11)
\end{tikzpicture}
    \caption{An unrestricted BT emulating a pure FSM}
    \label{fig:fsm_bt}
\end{figure}
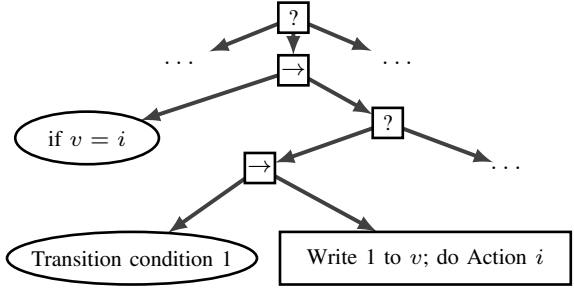
\begin{lem}~\cite{ogren2012increasing} \label{implict memory+bt=fsm}
\textbf{pFSM} $\subseteq$ \textbf{uBT}.
\end{lem}
\begin{IEEEproof}
Construct the BT shown in Fig.~\ref{fig:fsm_bt}, where the states of the FSM are numbered from 1 to $n$ and the current state is stored in the auxiliary variable $v$ (assume $v$ starts at a fixed value, say 1, and assign that to the start state of the FSM). At each step, check the state by checking $v$, then update $v$ according to the transition triggered in the FSM (which are in $\Inp$), while doing the action corresponding to that state in the FSM.
\end{IEEEproof}
\begin{thm}[The ASM Hierarchy] \label{asm hierarchy}
\begin{align*}
\text{\textbf{pTR}} = \text{\textbf{pDT}} = \text{\textbf{pBT}} &= \text{\textbf{Reactive}} \subset \\ \text{\textbf{uTR}} = \text{\textbf{uDT}} = \text{\textbf{uBT}} &= \text{\textbf{pFSM}} = \text{\textbf{uFSM}}
\end{align*}
\end{thm}
\begin{IEEEproof}
First observe that if we can prove that given an architecture of type $X$ we can construct an equivalent one of type $Z$, and this proof does not depend on whether the architectures are unrestricted or pure, then this suffices to prove both $\text{\textbf{pX}}\subseteq\text{\textbf{pZ}}$ and $\text{\textbf{uX}}\subseteq\text{\textbf{uZ}}$. Now, given any reactive ASM derived from a map $r:\Inp\to\Act$ or ($r:\Inp\times P\to\Act\times Y$), we can construct a FSM with the same ASM by constructing a complete directed graph where every arc is labelled by an input $x\in\Inp$, and goes from a node $q$ to the node $r(x)$. Thus \textbf{pFSM} is more expressive than the other pure architectures and \textbf{uFSM} is more expressive than the unrestricted ones. One can observe that DTs are at least as expressive as TRs, by constructing a tree of the form `\textbf{if} $k_1$ \textbf{select} $a_1$ \textbf{else} (\textbf{if} $k_2$ \textbf{select} $a_2$ \textbf{else} ($\dots$))', which has the same ASM as a TR $k_1\to a_1,\; k_2\to a_2,\; \dots$. Similarly, in \cite{generalise} the authors show how to construct a BT with the same ASM from a given DT (using different terminology).\footnote{The technical difficulty of this comparison that DTs do not take into account return values of Actions while BTs do. If Actions are assumed not to have return values, as is done in~\cite{generalise}, it is easy to see that they are identical. Even without this assumption it can be proved using the `Negation' Decorator node.} 
 Observe that ASMs from pure BTs, pure DTs and pure TRs are all reactive. Now, $\text{\textbf{Reactive}} \neq \text{\textbf{pFSM}}$, because the FSM with two states and a single transition between them is not reactive if that transition is possible. \emph{Claim:} $\text{\textbf{Reactive}} = \text{\textbf{pTR}}$. We require a TR which selects the same actions as an arbitrary reactive architecture given any input sequence. Let $r:\Inp\to\Act$ be the function defining the reactive ASM we wish to emulate, and let $y_1,y_2,\dots,y_m$ be the elements of $\Inp$. Then construct the TR
\begin{equation*}
    x = y_1 \to r(y_1),\ 
    \dots,\ 
    x = y_m \to r(y_m)
\end{equation*}
where $x$ is the last element of the input sequence in $\Inp^*$. This has the same ASM as $r$. \emph{Claim}: \textbf{pFSM} = \textbf{uFSM}. We omit the details, but it is a fairly straightforward proof to show that from an unrestricted FSM with states $Q$ and $m$ values for the auxiliary variable $v$, we can construct an equivalent pure FSM whose state set is $Q\times\{1,\dots,m\}$.
Then, Lemma~\ref{implict memory+bt=fsm} completes the proof.\footnote{This hierarchy can be summarised by the statement that sequential logic (FSMs) is not the same as combinational logic (reactive ASMs).}
\end{IEEEproof}


\subsection{Expressiveness versus readability} \label{sec:bts vs fsms}

In this section we discuss further the relationship between expressiveness and readability, by again comparing BTs and FSMs. In the previous section, the ASM Hierarchy informed us that FSMs are more expressive than pure BTs, but as expressive as unrestricted BTs. This seems to suggest that there is little point considering pure BTs, because their expressiveness is limited. However, pure BTs are worth thinking about because the arguments for why BTs are readable---and thus are useful tools---do not necessarily extend to unrestricted BTs. What we show now is that there is a trade-off between readability and reactiveness, in that pure BTs are less expressive but unrestricted BTs can be less readable. 

Fig.~\ref{fig:fsm_bt} depicts an unrestricted BT equivalent to a pure FSM whose structure does not resemble the behavior it induces. First, the order of the subtrees is irrelevant, so the visual information of order is unused. Secondly, contrary to one of the key benefits of BTs, considering each subtree in isolation does not allow one to understand the behavior. Finally, the structure of each individual selection does not convey which \emph{sequences} of selections are possible, as it is not reactive. If we wished to display visually the structure of this decision-making process, we would have a closer correspondence with its behaviors if we used the FSM in its original form, where the paths through the structure are explicit. Recall Dijkstra's words: ``we should do ... our utmost to shorten the conceptual gap between the static program and the dynamic process"~\cite{gotoharmful}. If an FSM behavior is desired, it seems that it would be more readable to represent it by a pure FSM explicitly, and not use an unrestricted BT. Unrestricted FSMs are also not very readable for this reason, but in Theorem~\ref{asm hierarchy} we showed that we do not lose any expressiveness by restricting ourselves to pure FSMs, so we would always use these.

It is useful to consider the analogous case of the B\"ohm-Jacopini Theorem from structured programming. This result, published in 1966, proved that all programs could be constructed using only \textbf{while} loops and \textbf{if} statements, if tests and assignments on some auxiliary variables were allowed. While this provided a formal grounding for structured programming, it also destroyed the `structure' of a program and so could not identify cases when a \textbf{go to} was in fact the clearest choice~\cite{knuth1974structured}.

This theorem was later shown to have an essentially trivial proof~\cite{cooper1967bohm}, where auxiliary variables store the state of the original program and a single enclosing \textbf{while} loop repeatedly selects the corresponding code blocks and updates the auxiliary program-state variables accordingly. This form of the proof exactly matches how the BT emulates the FSM in Lemma~\ref{implict memory+bt=fsm}. The repeated tick generation at the root corresponds to the outer \textbf{while} loop, with the auxiliary variable storing the FSM state and selecting the appropriate action. 

The B\"ohm-Jacopini theorem was widely cited by supporters of structured programming as an argument for the abolition of \textbf{go to}s. 
However, to quote Knuth~\cite{knuth1974structured}: ``from a practical standpoint [the B\"ohm-Jacopini Theorem] is meaningless ... we have eliminated all the \textbf{go to}s, but we've actually lost all the structure". Dijkstra, who cites this result as a theoretical justification for structured programming, himself points out that ``the exercise to translate an arbitrary flow diagram more or less mechanically into a jumpless one, however, is not to be recommended. Then the resulting flow diagram cannot be expected to be more transparent than the original one"~\cite{gotoharmful}. Indeed, in their original paper, B\"ohm and Jacopini conjectured, and Floyd and Knuth later proved~\cite{knuth1971notes}, that the auxiliary variables form a critical part of this proof---there are some programs which cannot be `structured' without additional variables.

Lemma~\ref{implict memory+bt=fsm}, originally proved in~\cite{ogren2012increasing} and reproved in~\cite{generalise,unifiedframework,btbook}, has similarly been used as a theoretical justification for structuring control architectures as BTs. What we have proved here then plays the part of~\cite{knuth1971notes}, as we have shown that accessing an auxiliary variable is critical in this proof---we cannot take an arbitrary FSM and produce an equivalent pure BT---that is, a BT using \emph{only} the given actions and transition conditions.
This `mechanical' approach to translating FSMs into unrestricted BTs cannot be expected to make them any more readable. Instead, it seems to make them less readable. In the next two sections we show how this trade-off exists in the `intermediate' cases used in practice between pure BTs and unrestricted BTs\footnote{The proof used to show that pure TRs are as expressive as any Reactive architecture likewise involved `deconstructing' the architecture, and so similarly we may not expect the output of that transformation to be more readable than the original.}.

\subsection{Computation in conditions} \label{sec:conditions}

In the previous section, we constructed unrestricted architectures using `special' actions and conditions to read and write from an auxiliary variable. This was required to make BTs as expressive as FSMs. Importantly, we did \emph{not} treat these actions as black boxes. The proof depended on the fact that after writing to the variable, its value is retained on our next reading. When all actions and conditions are treated as black boxes, we have only pure architectures.

In practice the meaning of actions and conditions does matter, and may influence the operation of the architecture, even without explicit read/write actions as in an unrestricted BT. An example is the following TR, which emulates an FSM: `if FSM would do A $\rightarrow$ do A; if FSM would do B $\rightarrow$ do B; etc'. The condition `FSM would do A' depends on past inputs to the structure. This is not readable, as its visual structure conveys no information about the sequences of actions it induces.

By the ASM Hierarchy, if a BT emulates an FSM, it must use conditions which depend on the past. However, we argue that such conditions reduce the readability of a BT, and so the trade-off between expressiveness and readability continues to apply. FSMs have a current state, so some information of their past decisions is visually stored in the structure. BTs, by contrast, have no representation of past decisions in their structure. In a pure BT, each input is essentially independent, so this isn't needed. If `past-dependent' conditions like `FSM would do A' are in $\Inp$, then formally they serve to forbid certain input sequences in $\Inp^*$. For example, if $x\in \Inp$ requires that $y$ has occurred in an earlier step, then any sequences in $\Inp^*$ with $x$ not preceded by $y$ are `impossible'. This information is not available in the BT structure however. Consider some concrete examples:

\begin{align}
    \text{if have opened door}&\seq \text{ walk through}\\
    \text{if `mode' is Z}&\seq \text{ do W} \\
    \text{if region X is unexplored}&\seq \text{ do Y}
\end{align}

In example (1), when a mobile robot is presented with an open door it is impossible to induce from the BT structure whether the robot walks through that door, without knowing if \emph{the robot itself} had opened it. 
In this particular case, a present-tense condition such as `if door open' may be more readable, as it removes the dependence on the past. The readability of example (2) depends on whether `mode' is influenced by past states. If `mode' is a physical button on the robot, then maybe this is readable; if `mode' is an FSM state as in Fig.~\ref{fig:fsm_bt} then perhaps less so, because this is the case of Section~\ref{sec:bts vs fsms}. In such a case, it might be more readable to separately provide the FSM controlling `mode', then have each `mode' controlled by a BT. Likewise in (3), `region X unexplored' depends on where the robot has explored, which is not clear from the structure. However, this could be more readable if a representation of the robot's internal `map' is available.

We postulate that using conditions that depend in more complex ways on past states makes BTs less readable. However, dependence on the past is required for more expressiveness, by Theorem~\ref{asm hierarchy}, showing there is a trade-off between these in BT design. These must therefore be carefully considered and balanced.

\subsection{Do memory nodes increase expressiveness?} \label{sec:addingmemory}

We have seen that pure BTs are less expressive than FSMs. However there exist extensions to BTs which could be used to overcome these limitations on expressiveness, possibly in a more readable way than using auxiliary variables. For instance, `Control flow nodes with memory' and Decorator nodes implementing what we call `eternal memory' are both sometimes used as simple extensions which use memory in BT decision-making, but do so in a way that is explicitly represented in the structure. Do these additions allow us to increase the expressiveness of pure BTs? The answer is yes, but not to the level of an unrestricted BT---pure BTs with these extensions are still strictly less expressive than FSMs. 
Firstly, we explain these concepts.

\emph{Sequence and Fallback with memory}~\cite{btbook}, written $\seq^*$ and $\fb^*$, are variants on the Sequence and Fallback control flow nodes which prevent unnecessary ticking in circumstances where reactiveness is not desired. The Sequence with memory ticks its children left to right, but remembers which have already returned Success, and so does not tick them again at each time step. If any child returns Running or Failure the Sequence with memory also returns that value. If the Sequence with memory returns Success or Failure, then the memory is `reset', and on the next received tick it begins at the leftmost node.  Such nodes are generally suggested for circumstances where reactiveness is not needed, such as when it is known that the leftmost child would continue to return Success after succeeding~\cite{btbook}. It can be seen that this is not reactive, but retains the tree structure and interface of BTs. The Fallback node with memory operates similarly, with the role of Success and Failure interchanged.

\emph{Run until Success (Failure)} or \emph{eternal memory:} This Decorator node ticks its one child and returns that child's return value until the child returns Success, after which the Decorator returns Success always. It is easy to see that as its behavior is dependent on past return values, it is not reactive. However, its simplicity keeps it readable. We refer to this Decorator by the name \emph{eternal memory} to contrast it with the `periodic resets' of control flow nodes with memory.

\begin{figure}
    \centering
    \begin{tikzpicture}[shorten >=1pt,node distance=2cm,on grid,auto]
    \node[initial,state] (A) {$A$};
    \node[state] (B) [right=of A] {$B$};
    \node[state] (C) [right=of B] {$C$};
    \path[->]
        (A) edge node {$s$} (B);
    \path[->]
        (B) edge[bend right=20] node {$s$} (C);
    \path[->]
        (C) edge[bend right=20] node[swap] {$s$} (B);
    \path[->]
        (C) edge[bend right=45] node[swap] {$f$} (A);
    \end{tikzpicture}
    \caption{An FSM $M$, whose ASM cannot be constructed by a BT using memory control flow nodes and decorators.}
    \label{fig:fsm counterexample}
\end{figure}
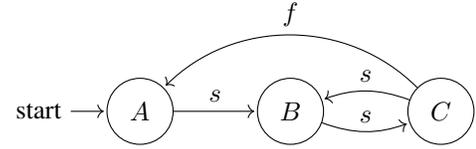

\begin{thm}\label{thm:mbt_hier}
Let \textbf{mBT} be the set of ASMs which are derived from pure BTs with eternal memory and control flow nodes with memory allowed. Then $\text{\textbf{pBT}}\subset \text{\textbf{mBT}}\subset\text{\textbf{uBT}}$.
\end{thm}
\begin{IEEEproof}
It is easy to see that $\text{\textbf{pBT}}\neq \text{\textbf{mBT}}$, for instance by considering the behavior `do $A$ until it returns Success, then do $B$ forever'. This is straightforward to construct with an eternal memory Decorator. However, in a pure BT if on the first input $x$, $A$ returns running then $A$ is selected and $A$ must also be selected in any later occurrence of $x$, even if $A$ returned Success in between. Now we show that $\text{\textbf{mBT}}\subset\text{\textbf{FSM}}=\text{\textbf{uBT}}$. Consider the FSM $M$ shown in Fig.~\ref{fig:fsm counterexample}. Let $T$ be some pure BT, but which may contain control flow nodes with memory and eternal memory Decorators. We show that no such $T$ can have the same ASM as $M$. Firstly, note that by our assumption that FSMs make only finitely many transitions on a given input, there cannot be any inputs for which both $B$ and $C$ return Success, and no input for which $A$ and $B$ return Success and $C$ returns Failure, as in these cases the FSM does not make a selection. Otherwise, we assume all other combinations of return values of $A$, $B$ and $C$ are inputs. We write inputs $x\in\Inp$ as $x=(v_A,v_B,v_C)$ where $v_A$, $v_B$ and $v_C$ are the return values in $x$ of $A$, $B$ and $C$. We write $s$, $f$ and $r$ for these taking the values Success, Failure and Running respectively. As $M$ depends solely on these variables, any other propositions are irrelevant for any $T$ which emulates $M$. The transitions labelled $s$ and $f$ are assumed to be triggered by any input where the action they emerge from return Success or Failure respectively. This makes comparison with BTs easier.

Consider the input sequence \[
z = (r,r,r),(s,r,r),(r,s,r),(r,f,s),(r,s,r),(r,r,f)
\] to which $M$ selects $y = A,B,C,B,C,A$. Suppose $T$ contains eternal memory. Observe that once a subtree rooted by an eternal memory Decorator return Success or Failure, it is never ticked again. Suppose we apply the input sequence $z^n$ to $M$, which is $z$ repeated $n$ times. Suppose the actions selected by $T$ on the $n$th copy of $z$ are identical to the selections for the $(n-1)$-th copy. Then, as every eternal memory Decorator returns the same value, the output of $T$ must be an identical pattern of six actions for all subsequent copies of $z$. Thus eventually, after some large enough $n$, all eternal memory subtrees either always return Success or Failure or are never ticked in subsequent inputs of the sequence $z$. Thus, in the last six inputs of $z^{n+1}$, $T$ must behave as if it contained no eternal memory. We now focus on these last inputs. Suppose $T$ matches the first four outputs of this sequence. However, in the fourth input $(r,f,s)$, $T$ must select $B$, which returns Failure in that state. As a result the top-level tree returns Failure, and thus all control flow nodes with memory are reset. On the subsequent input $(r,s,r)$ $T$ must select $A$, as pure BTs without memory nodes always select their leftmost node if possible, and this node must be $A$ as $(r,r,r)$ selects $A$. This differs from $M$, so $T$ cannot realise the same ASM.
\end{IEEEproof}
Thus, these BT extensions represent possible trade-offs between readability and expressiveness. They are more expressive than pure BTs, though arguably less readable. Conversely, they are arguably more readable than unrestricted BTs, but also less expressive. It is not clear which real-world tasks this class of expressiveness might correspond to. This result does not rule out the possibility of other BT extensions increasing expressiveness to the level of FSMs, but the results of this paper suggest that such constructions are unlikely to be more readable than FSMs themselves.

The following, a consequence of Theorems \ref{asm hierarchy} and \ref{thm:mbt_hier}, summarises the main result of the paper regarding the expressiveness of BTs and other control architectures: 
\begin{corol}
\begin{align*}
\text{\textbf{pTR}} = \text{\textbf{pDT}} = \text{\textbf{pBT}} &= \text{\textbf{Reactive}} \subset \text{\textbf{mBT}} \\ \subset \text{\textbf{uTR}} = \text{\textbf{uDT}} = \text{\textbf{uBT}} &= \text{\textbf{pFSM}} = \text{\textbf{uFSM}}.
\end{align*}
\end{corol}

\section{Conclusions and Future Work} \label{sec:conclusions}
In this paper we compared the expressiveness of BTs and other control architectures. We showed there exists a trade-off between readability and expressiveness, which indicates that BT design requires careful thought. Future work could include richer comparisons based on these ideas, and drawing further inspiration for AI design from programming paradigms like structured programming. 

\bibliographystyle{IEEEtran}
\bibliography{references}

\end{document}